\pdfoutput=1
\documentclass{article}

\PassOptionsToPackage{numbers, compress}{natbib}
\usepackage[preprint]{neurips_2026}

\usepackage[utf8]{inputenc} 
\usepackage[T1]{fontenc}    
\usepackage{hyperref}       
\usepackage{url}            
\usepackage{booktabs}       
\usepackage{amsfonts}       
\usepackage{nicefrac}       
\usepackage{microtype}      
\usepackage{xcolor}         
\usepackage{graphicx}
\usepackage{amsmath}
\usepackage{placeins}

\title{A Spiking Ring-Attractor Architecture for Representation and Control of Revolute Robot Joints}

\author{%
  Aung Htet \quad Alejandro Rodriguez Jimenez \quad Ruidong Ma \quad Alessandro Di Nuovo \\
  Sheffield Hallam University, Sheffield, UK \\
  \texttt{c3069061@hallam.shu.ac.uk}, \texttt{A.Jimenez-Rodriguez@shu.ac.uk}, \\
  \texttt{Ruidong.Ma@shu.ac.uk}, \texttt{a.dinuovo@shu.ac.uk}
}

\begin{document}

\maketitle

\begin{abstract}
Robotic systems with revolute joints evolve on circular state spaces, yet most control architectures represent these states using Euclidean variables that ignore this underlying geometry. We propose a fully spiking neural control architecture in which ring attractors encode joint states, enforcing representations consistent with the topology of revolute systems.

Homeostatic regulation computes goal-state mismatch directly over these distributed representations, and opponent gain-modulated populations convert this mismatch into controlled phase shifts of the activity bump, implementing set-point regulation as motion on the ring.

We show that the bump dynamics follow an $SO(2)$ geometry, and that gain asymmetries induce a signed phase velocity on the ring. To interface distributed activity with scalar regulation signals, we derive an analytic Fourier phase decoder and implement its signed components using push--pull spiking populations.

For multi-joint systems, we demonstrate that local ring-based Fourier features compose into product representations that linearly decode $SO(3)$ orientations through forward kinematics. Experiments in NEST validate the full $SO(2)$ regulation mechanism and the spiking product-feature decoder for lift, pitch, and yaw. Robot experiments on a Tiago arm show that decoded ring dynamics generate joint velocity commands, producing directed motion with joint-dependent accuracy.

Together, these results establish a neuromorphic control framework in which representation, regulation, and action are unified within a ring-attractor substrate aligned with the geometry of revolute systems.

\end{abstract}

\section{Introduction}

Neural circuits in biological systems often do not appear to separate representation from control. Internal states are maintained as distributed activity patterns, goals can be encoded within the same substrate, and behavior can emerge through continuous regulation of one activity state toward another. This organization is especially clear in neural systems for orientation and navigation, where heading, goal direction, and steering commands are represented and transformed within population activity \citep{green2017,westeinde2024,pires2024}. Ring attractors provide a particularly relevant example: in biological head-direction systems, they maintain localized activity bumps on a circular manifold and support angular integration \citep{seelig2015,evans2017,kim2017}. This circular structure has also made ring attractors useful in computational and robotic models of path integration, grid-cell organization, and bio-inspired navigation \citep{knowles2023,pantelis2022,rivero2023,benas2024}.

Building on this circular population structure, we propose a robotic control
architecture in which ring attractors represent revolute joint states and serve
as dynamical substrates for control. The central idea is to regulate the phase of
a persistent activity bump directly, rather than decoding joint state into an
external Euclidean variable before control. Homeostatic comparison provides a
signed mismatch between goal and state, while opponent gain modulation converts
that mismatch into directed transport of the bump around the ring
\cite{boulant2006neuronal,salinas2001,zhang2000}. Because the neural activity
bump and the physical joint angle both evolve on circular manifolds, this yields
a topology-preserving, fully spiking control primitive for revolute variables.

This perspective also provides a natural geometric and compositional structure.
Phase shifts of a ring-attractor bump correspond to motion on \(SO(2)\), with
infinitesimal displacements described by a Lie-algebraic tangent direction
\cite{murray2017,wu2024,khona2022}. To interface this distributed representation
with scalar regulatory signals, we derive an NEF-inspired Fourier decoder that
extracts phase from ring activity \cite{eliasmith2005unified}. For multi-joint
systems, the same local Fourier features, \(\cos q_i\) and \(\sin q_i\), compose
through known forward kinematics to provide an \(SO(3)\) orientation readout.
Thus, individual rings generate controlled \(SO(2)\) trajectories, while their
decoded features support robot-level orientation estimates.

Our contributions are: (i) a biologically inspired closed-loop ring-attractor
controller in which a Hammel-style homeostatic comparator drives opponent gain
modulation to transport bump states on an \(SO(2)\)-structured manifold; (ii) a
Lie-algebraic transport analysis of the gain-imbalance mechanism; (iii) an
analytic Fourier phase decoder that maps distributed ring activity to scalar
phase estimates compatible with homeostatic comparison; (iv) a forward-kinematic
Fourier decoder that reads out an \(SO(3)\) orientation estimate from known
revolute-chain ring states via product Fourier features; and (v) empirical
validation in NEST spiking simulations \cite{gewaltig:NEST} and on a Tiago robot
\cite{Pags2016TIAGoTM}.

The remainder of the paper presents the architecture, derives the \(SO(2)\)
transport mechanism and Fourier readouts, and evaluates the resulting spiking
controller in NEST and on a Tiago arm. Additional derivations and experimental
details are provided in the appendix.

\section{Architecture Overview}

\subsection{Single-ring architecture for revolute joint control}

\begin{figure}[!htbp]
    \centering
    \includegraphics[width=\linewidth]{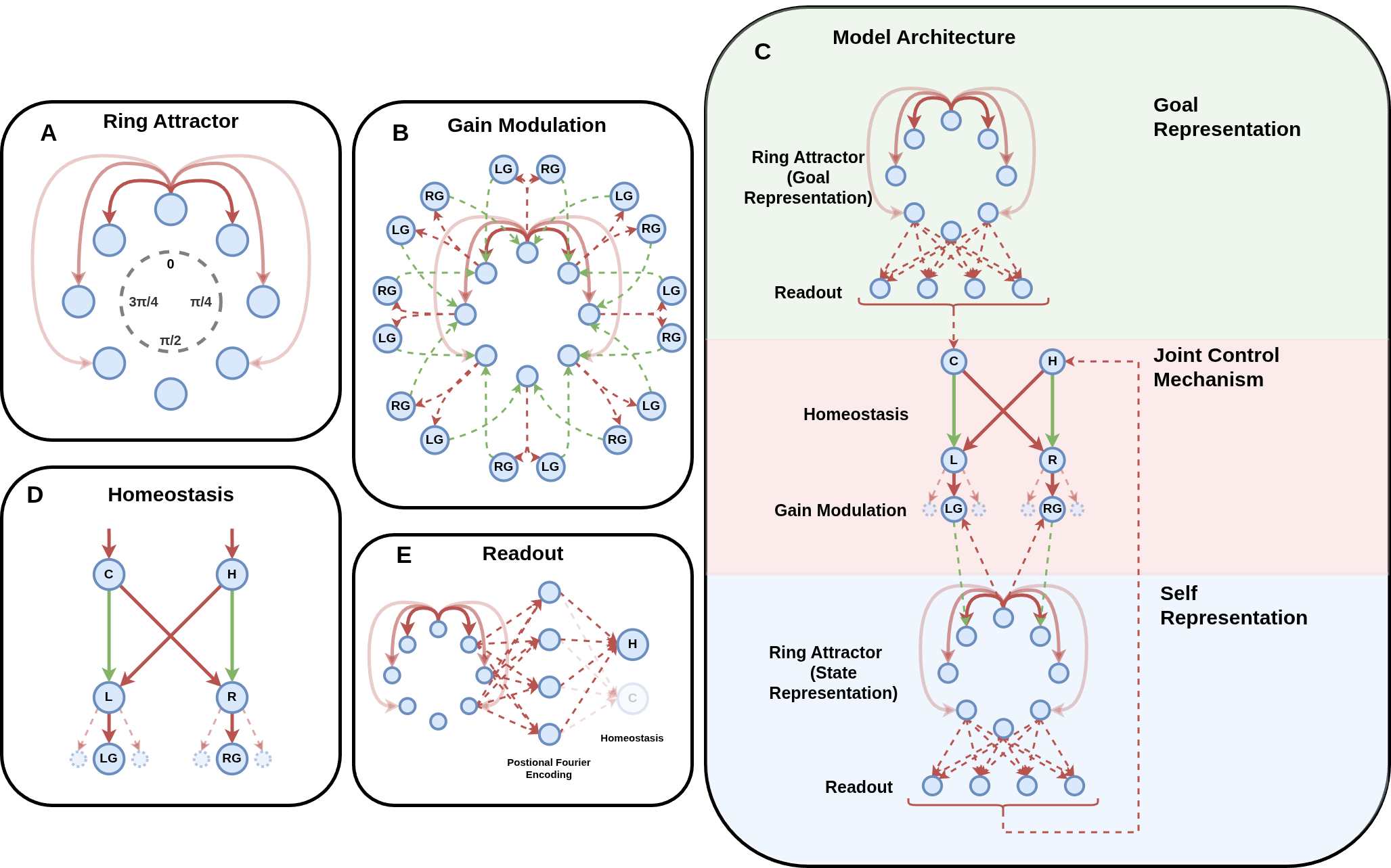}
    \caption{
    Single-ring architecture for revolute joint control, composed of four modules.
    The goal ring encodes the desired joint phase as a localized bump; the state ring encodes the current joint phase.
    A Positional Fourier readout decodes each bump into a scalar phase estimate, which feeds a Hammel-style homeostatic comparator.
    The comparator drives opponent left-gain (LG) and right-gain (RG) populations, moving the ring in their preferred direction.
    A mismatch between goal and state produces a gain imbalance that moves the bump state along the ring; when the mismatch vanishes, the opponent signals become balanced and stabilizes the bump.
    }
    \label{fig:model_architecture}
\end{figure}

This single-ring architecture builds on prior ring-attractor work for number subitization \cite{htet2025}, extending the substrate from discretized representations of continuous revolute-joint control. Figure~\ref{fig:model_architecture} shows the resulting control primitive for a single revolute joint, composed of goal and state ring-attractor populations, a Fourier readout layer, a homeostatic comparator, and opponent gain-modulation populations.

The goal ring encodes the desired joint configuration as the phase of a localized activity bump, and a similar representation is subsequently used by the state ring for the current joint configuration. The activity of these rings is projected onto positional Fourier features, which are decoded into spike-rate estimates of scalar phase. These goal and state phase estimates are then provided to a Hammel-inspired homeostatic comparator for feedback control \cite{boulant2006neuronal}. The comparator produces opponent output signals that drive the left-gain (LG) and right-gain (RG) populations, directing the state bump along the preferred angular direction and closing the control loop.

This single-ring unit constitutes the basic control primitive. The following sections first derive the geometry and transport dynamics of this primitive, before Section~2.5 describes how multiple ring states are decoded through a known forward-kinematic chain.

\subsection{Ring attractor as an \(SO(2)\)-structured state space (Figure~\ref{fig:model_architecture}.A)}

A ring attractor is a recurrent neural network whose connectivity is invariant
to rotations of a circular preferred-phase index
\citep{zhang1996,amari1977,gerstner2014}. This symmetry supports a continuous
family of stable population states: localized activity bumps centered at any
phase \(\phi \in S^1\). In the regime considered here, the controlled state is
the bump centroid rather than the activity of any individual neuron.

For a revolute joint, this representation is natural because the joint angle and
the bump phase both live on a circular manifold. We approximate the ring activity
as a translated bump,
\begin{equation}
    u_r(\theta,t) \approx f\!\left(\theta - \phi(t)\right),
    \label{eq:bump_approx}
\end{equation}
where \(f\) is the bump profile and \(\phi(t)\) is the represented joint phase.
Thus, although the neural activity is high dimensional, the controlled coordinate
is the scalar phase \(\phi(t)\in S^1\).

This population code has an \(SO(2)\) structure: shifting the bump by an angle
\(\alpha\) maps \(u_\phi\) to \(u_{\phi+\alpha}\). The family of bump states is
therefore an orbit of the rotation group, and the centroid phase can equivalently
be represented as \(R(\phi)\in SO(2)\). Control corresponds to moving the bump
along this orbit. The tangent direction is generated by the infinitesimal
rotation operator \(J=-\partial_\theta\), so an input that reliably changes the
represented joint angle must project onto the bump tangent rather than merely
perturbing firing rates.

In a symmetric ring, all centroid phases are equally stable and there is no
preferred direction of drift. Balanced afferent drive preserves this symmetry and
keeps the bump stationary. Directed motion requires breaking the symmetry through
a signed imbalance between opponent directional drives, which produces a
transport term of the form
\begin{equation}
    \partial_t u_r(\theta,t)+c(u_r)\partial_\theta u_r(\theta,t)=F_r(u_r),
    \label{eq:transport}
\end{equation}
where \(F_r(u_r)\) denotes the symmetric recurrent dynamics and \(c(u_r)\) is a
signed transport coefficient. The next section shows how opponent gain modulation
implements this coefficient in the spiking circuit.

\subsection{Gain modulation as Lie-algebraic transport (Figure~\ref{fig:model_architecture}.B)}

Opponent gain modulation provides the directional drive that moves the bump. The
architecture uses two gain populations, \(u_g^+(\theta,t)\) and
\(u_g^-(\theta,t)\), which receive phase-shifted copies of the state-ring
activity together with scalar homeostatic drives:
\begin{equation}
    u_g^+(\theta,t)
    =
    W_{rg}\, g(u_r(\theta+\delta\theta,t)) + B_+(t),
    \qquad
    u_g^-(\theta,t)
    =
    W_{rg}\, g(u_r(\theta-\delta\theta,t)) + B_-(t).
    \label{eq:gain_pops}
\end{equation}
The opposite shifts make the two populations sensitive to opposite directions of
bump motion. When \(B_+(t)=B_-(t)\), their effects are balanced and the ring
symmetry is preserved. When the homeostatic comparator creates an imbalance, the
symmetry is broken and the bump drifts with a sign determined by
\(B_+(t)-B_-(t)\).

To first order in the phase shift \(\delta\theta\), the summed gain feedback
contains a symmetric component and a directional component proportional to the
ring gradient:
\begin{equation}
    g(u_g^+) + g(u_g^-)
    \approx
    \underbrace{g(x_+) + g(x_-)}_{\text{symmetric}}
    +
    \underbrace{W_{rg}\delta\theta\, g'(u_r)
    \bigl[g'(x_+) - g'(x_-)\bigr]
    \partial_\theta u_r}_{\text{directional}},
    \label{eq:opponent_sum}
\end{equation}
where \(x_\pm = W_{rg}g(u_r)+B_\pm(t)\). The symmetric term changes the net drive
without selecting a direction. The directional term vanishes when the opponent
drives are balanced and becomes nonzero only when \(B_+(t)\neq B_-(t)\).

Collecting this directional feedback in the ring dynamics yields
\begin{equation}
    \partial_t u_r(\theta,t)
    +
    c(u_r)\,\partial_\theta u_r(\theta,t)
    =
    F_r(u_r),
    \label{eq:transport_ring}
\end{equation}
with transport coefficient
\begin{equation}
    c(u_r)
    =
    \frac{W_{gr}W_{rg}\,\delta\theta}{\tau_r}
    \,g'(u_r)
    \bigl[g'(x_+) - g'(x_-)\bigr].
    \label{eq:transport_coeff}
\end{equation}
Thus the transport speed scales with the phase shift, bidirectional coupling, and
opponent gain difference. Since \(J=-\partial_\theta\), this is equivalently
activity injection along the tangent direction of the \(SO(2)\) bump manifold:
\begin{equation}
    \partial_t u_r(\theta,t)
    =
    F_r(u_r) + c(u_r)\,J u_r(\theta,t).
    \label{eq:lie_transport}
\end{equation}

On the single-bump manifold, this transport reduces to an effective phase
velocity,
\[
    \dot{\phi}(t)=c_{\mathrm{eff}}(t),
\]
where \(c_{\mathrm{eff}}(t)\) is the projection of the gain imbalance onto the
bump tangent. Opponent gain modulation therefore encodes velocity along the ring,
not joint angle itself; integrating this velocity transports the internal joint
state toward the goal set by the homeostatic comparator.

\subsection{Positional Fourier readout for scalar phase decoding (Figure~\ref{fig:model_architecture}.E)}

The homeostatic comparator operates on scalar variables, whereas the ring
attractor represents state as a distributed bump. We therefore decode each ring
activity profile into a scalar phase estimate using a periodic Fourier readout.

For a translated bump \(g(x)=r(x-\theta)\), the centroid estimator can be written
as a projection of activity onto a positional ramp. Since the ring is periodic,
this ramp admits a Fourier sine expansion, yielding the linear decoder
\begin{equation}
    \hat{\theta}
    =
    b + \sum_{m=1}^{K} w_m S_m(g),
    \qquad
    S_m(g)
    =
    \sum_{x=0}^{N}
    g(x)\sin\!\left(\frac{2\pi m x}{L}\right),
    \label{eq:fourier_decoder}
\end{equation}
where \(w_m\) are analytic coefficients, optionally corrected for finite bump
width. Appendix~C gives the full derivation.

\begin{figure}[t]
    \centering
    \includegraphics[width=0.9\linewidth]{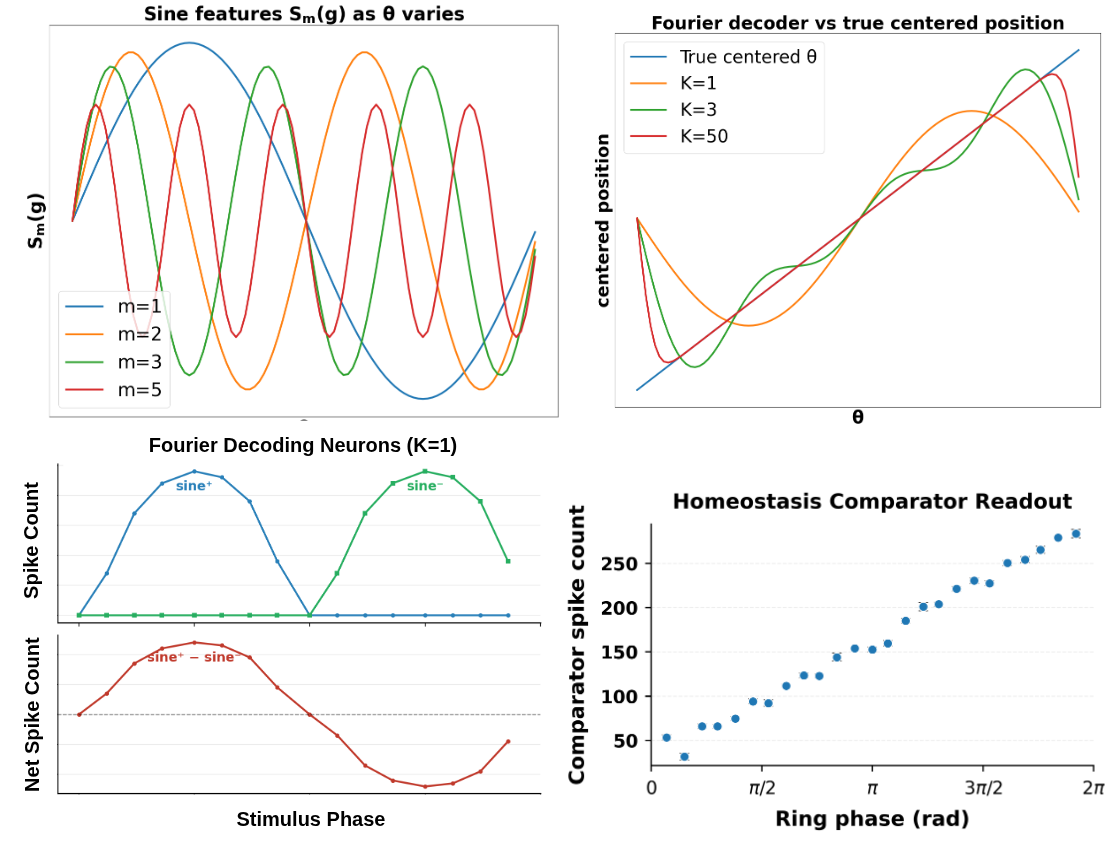}
    \caption{
    Analysis of the Fourier phase decoder.
    Top left: sine features \(S_m(g)\) for Gaussian bump activity.
    Top right: Fourier decoders with \(K=1,3,50\) modes reconstruct centered
    position with increasing accuracy.
    Bottom left: push--pull spiking implementation of the \(K=1\) feature.
    Bottom right: mean \(\pm\) standard deviation spike response across
    10 repeats and 25 ring positions, showing a stable phase-dependent readout.
    }
    \label{fig:combined_decoder_analysis}
\end{figure}

Figure~\ref{fig:combined_decoder_analysis} validates the decoder on translated
Gaussian bumps and its spiking implementation. Increasing the number of modes
improves phase reconstruction, while a single mode already captures the coarse
phase signal. Because the features \(S_m(g)\) are signed but spike counts are
nonnegative, each feature is implemented with push--pull subpopulations and
recovered by subtraction.

In the controller, the goal and state rings are decoded into
\(\hat{\theta}_g\) and \(\hat{\theta}_s\). The Hammel comparator computes their
signed mismatch and drives the opponent gain populations, thereby connecting the
\(SO(2)\)-structured population code to the gain-modulated transport mechanism.

\subsection{Forward-kinematic Fourier decoder for known revolute chains}

The single-ring controller represents and regulates one angular state. To connect
multiple controlled rings to robot-level variables, we define a
forward-kinematic readout for known revolute chains. This readout does not add a
new control law; it maps distributed ring activity into an end-effector
orientation estimate.

For each joint \(q_i\), the Fourier readout provides first-harmonic estimates
\[
    \bar{C}_i \approx \cos q_i,
    \qquad
    \bar{S}_i \approx \sin q_i,
\]
which form the local feature vector
\begin{equation}
    \varphi_i
    =
    \begin{bmatrix} 1 \\ \bar{C}_i \\ \bar{S}_i \end{bmatrix}
    \approx
    \begin{bmatrix} 1 \\ \cos q_i \\ \sin q_i \end{bmatrix}.
    \label{eq:joint_features}
\end{equation}
For a revolute joint with known axis, the corresponding rotation matrix is linear
in these features. For a serial chain, products of joint rotations therefore
generate products of local Fourier terms, all contained in
\begin{equation}
    \Phi^{(n)}
    =
    \varphi_1 \otimes \varphi_2 \otimes \cdots \otimes \varphi_n.
    \label{eq:chain_features}
\end{equation}
Thus, for a fixed kinematic convention, there exists a linear map \(T\) such that
\begin{equation}
    \mathrm{vec}(\hat{R}_{\mathrm{ee}}) = T\hat{\Phi}^{(n)},
    \label{eq:estimated_fk}
\end{equation}
where \(\hat{\Phi}^{(n)}\) is formed from ring-estimated features.

In the robot experiments, this decoder is used to read out lift, pitch, and yaw
under a fixed ordered convention. In the spiking implementation, signed
product-feature populations are formed from the input rings and projected to
orientation-coordinate output rings, allowing the composition to remain within
the neural substrate.

This decoder assumes known revolute kinematics, fixed rotation axes and
multiplication order, and reliable first-mode sine--cosine estimates from each
ring. Under these assumptions, it maps a product of local \(SO(2)\) ring states
to an \(SO(3)\) orientation estimate, extending the single-joint control
primitive toward task-space representations.

\section{Implementation and Results}
\subsection{Experimental setup}

\begin{figure}[t]
    \centering
    \includegraphics[width=\linewidth]{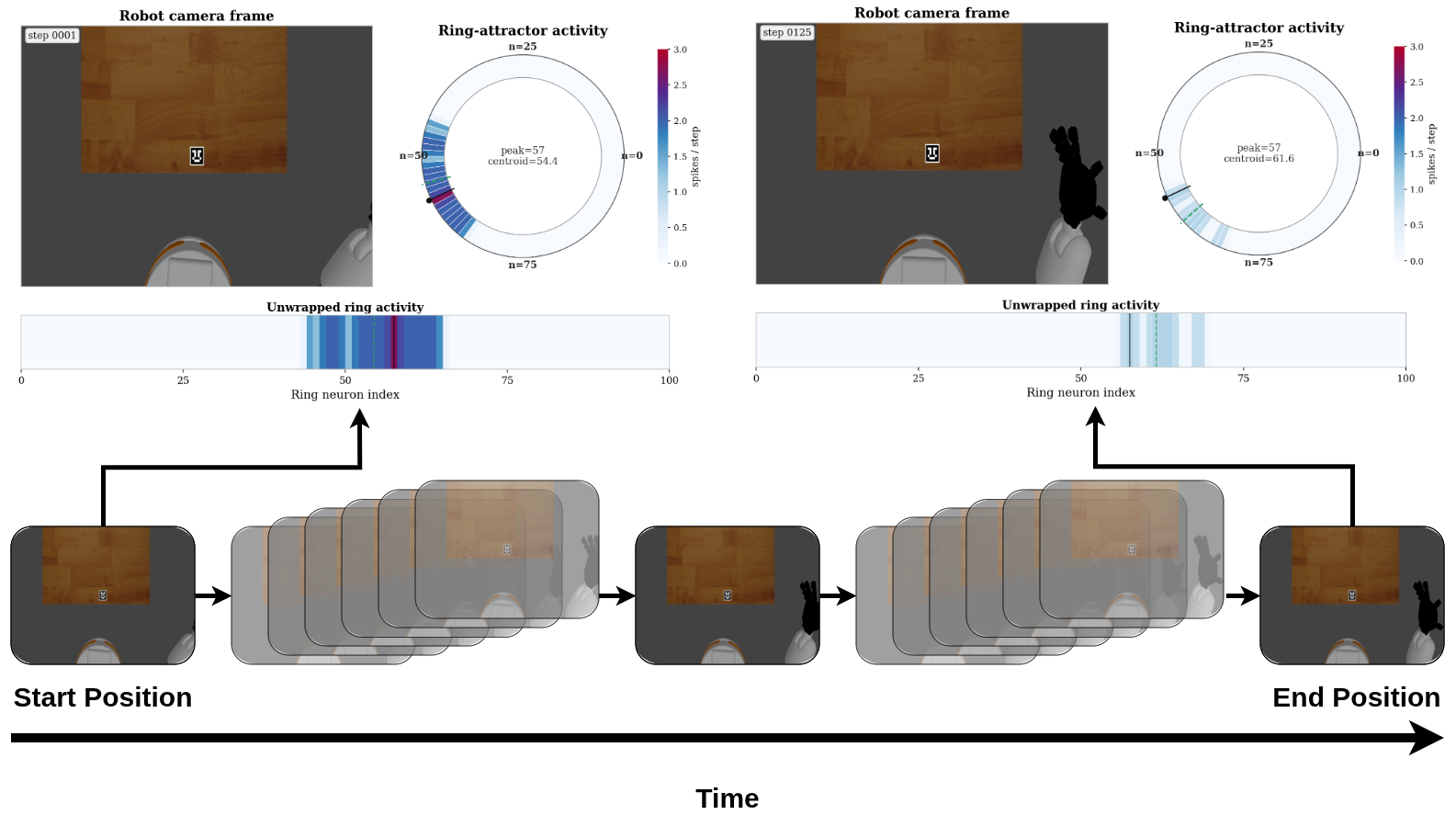}
    \caption{
    Robot-to-ring interface for the Tiago experiments. Each control step pairs a Tiago camera frame with the current spiking ring activity. The circular and unwrapped plots show the same population response, making the bump location and centroid visible during control. The timeline shows decoded gain-modulated commands moving the joint from start to target. Quantitative tracking results are reported in Figure~\ref{fig:tiago_single_joint_error}, with additional analyses in Figures~\ref{fig:single_joint_extra_info} and~\ref{fig:single_joint_extra_info_2}.
    }
    \label{fig:robot_ring_experiment_visualization}
\end{figure}

We evaluate the architecture in two stages. Spiking experiments instantiate the 
full architecture in NEST \cite{gewaltig:NEST} using \texttt{iaf\_psc\_alpha} populations and test 
whether the homeostatic gain-modulation loop transports a state bump toward a 
goal on the \(SO(2)\)-structured manifold, with model 
parameters reported in Appendix Table~\ref{tab:model_params}. Robot experiments connect the 
resulting trajectories to a Tiago arm \citep{Pags2016TIAGoTM} via ROS, using a receding-horizon 
streaming interface to issue joint velocity commands and logging measured joint 
states to evaluate whether decoded ring phases correspond to physical motion. 
For multi-ring trials, end-effector orientation is computed from logged joint 
angles through the known forward-kinematic model.

\subsection{Spiking implementation of \(SO(2)\) homeostatic transport}

We first evaluate the complete single-ring controller in NEST. The circuit
contains the state ring attractor, the Hammel-style homeostatic comparator, and
opponent left- and right-gain populations from Section~2. The goal is to verify
the core mechanism from Section~2.3: a phase mismatch should induce an opponent
gain imbalance, and this imbalance should transport the ring-attractor bump.

\begin{figure}[t]
    \centering
    \includegraphics[width=\linewidth]{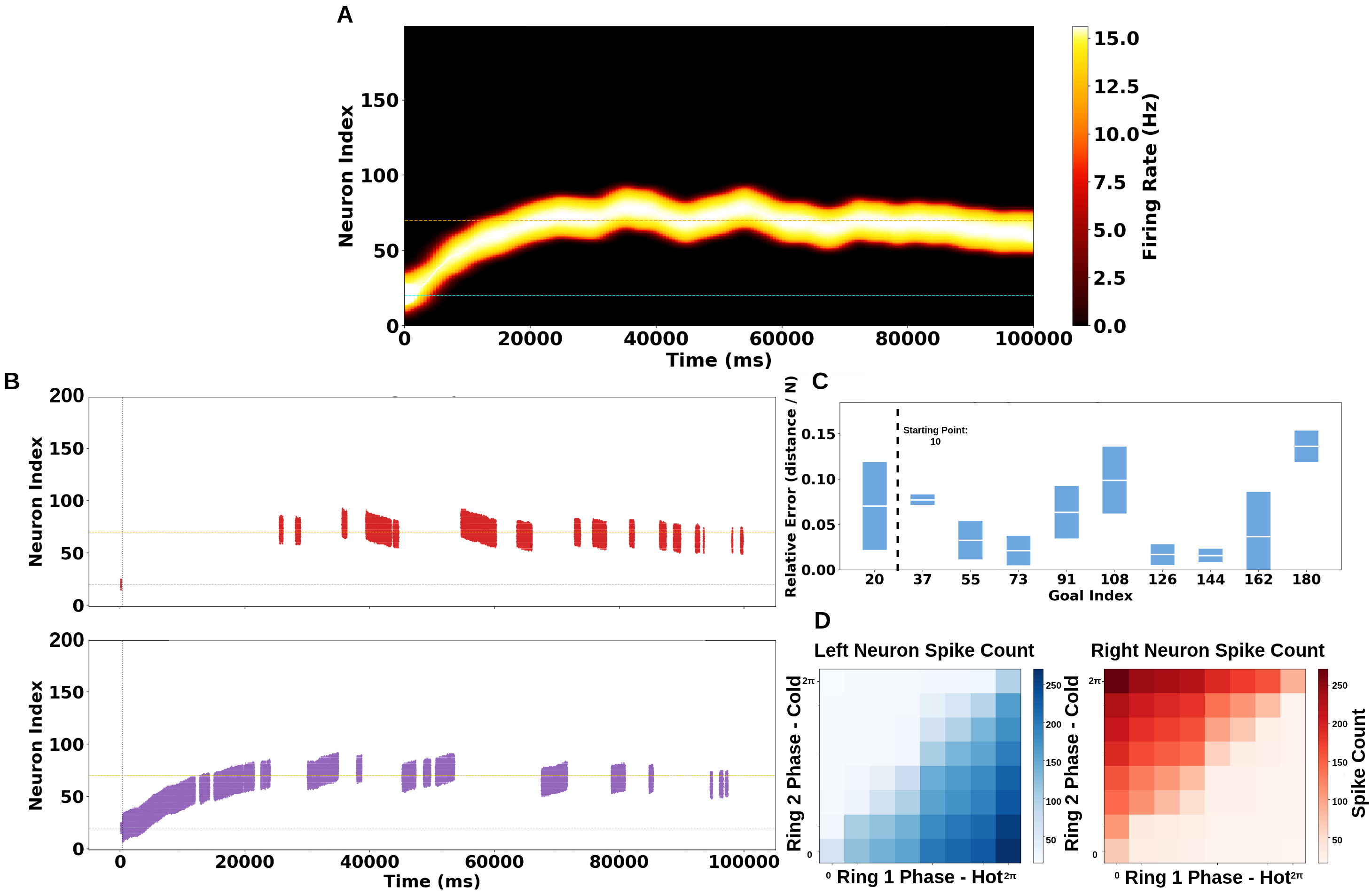}
    \caption{
    Single-ring \(SO(2)\) homeostatic controller implemented in NEST where the ring attractor is simulated using 200 neurons. 
    \textbf{A}: ring-attractor firing-rate heatmap with the decoded circular centroid 
    overlaid in white. The cyan dotted line marks the start phase and the orange 
    dotted line marks the goal phase. 
    \textbf{B}: left- and right-gain raster plots throughout the simulation period (associated to the top graph).
    \textbf{C}: relative goal error, computed as circular distance to the goal 
    divided by ring size \(N\), averaged over \(t>30000\,\mathrm{ms}\) across ten 
    runs per goal.  The black dotted vertical line marks the starting point.
    \textbf{D}: left- and right-homeostasis neuron spike-count heatmaps.
    }
    \label{fig:so2_spiking_architecture_analysis}
\end{figure}

Figure~\ref{fig:so2_spiking_architecture_analysis} shows the resulting activity. The ring
maintains a localized bump whose centroid is displaced from the start phase
toward the goal phase, demonstrating directed transport of the represented state.
The trajectory is not perfectly monotonic and does not always converge exactly
to the target, but it remains a coherent bump motion rather than a diffuse change
in population firing.

Panel~C quantifies the remaining circular error across target locations after
the transient period. Errors are target-dependent, indicating that the present
parameterization produces a usable but nonuniform transport field over the ring.
This nonuniformity is consistent with finite-size spiking effects, decoder
resolution, and local gain calibration, and motivates the calibration refinements
used in the robot interface.

Panel~D shows the predicted opponent structure in the homeostatic and gain
pathways. Hot and cold comparator activity varies systematically with relative
phase, producing a graded mismatch signal. The left- and right-gain channels show
complementary activation, with each channel preferentially engaged for one sign
of the phase mismatch. This push--pull pattern implements the imbalance
\(B_+(t)-B_-(t)\) that generates the transport coefficient in Equation~(12).
Together, these results show that the spiking circuit realizes the intended
symmetry-breaking mechanism for \(SO(2)\) bump transport, while also revealing
the target-dependent residual error expected from an unoptimized finite spiking
implementation.

\subsection{Tiago single-joint control from gain-modulated ring velocity}

We next evaluate whether the \(SO(2)\) ring controller can drive physical joint 
motion on the Tiago arm. The central test is whether opponent gain modulation 
produces a signed velocity command whose integration moves a revolute joint 
toward a target angle.

For each Tiago arm joint, the measured joint angle is mapped onto the state ring 
and a sampled target angle is injected into the goal ring. The NEST controller 
runs in closed loop; at each step, the difference in spike counts between the 
right-gain and left-gain populations,
\begin{equation}
    s_t = R_t - L_t,
    \label{eq:gain_signal}
\end{equation}
provides the signed gain signal. This is the spiking realisation of the gain 
imbalance derived in Section~2.3: positive \(s_t\) drives transport in one 
angular direction, negative \(s_t\) in the other. The signal is passed through 
an exponential filter and decoded into a joint velocity command with separate 
calibration gains for the positive and negative drive components, preserving the 
opponent structure of the controller. Commands are sent as receding-horizon 
joint trajectories via ROS, and measured joint positions are logged throughout.
\begin{figure}[t]
    \centering
    \includegraphics[width=0.8\linewidth]{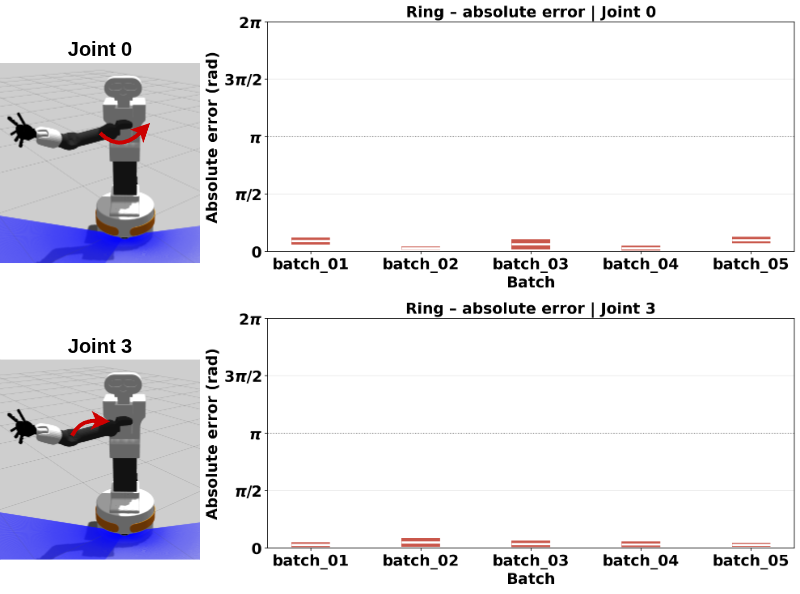}
    \caption{
    Tiago single-joint control where the opponent gain-modulation activity is translated to the robot's joint velocity.
    Left: Robot simulation snapshot demonstrating how the robot moves for specific joint index.
    Right: Robot simulation used to demonstrate error between goal state and target state of the ring attractor based single joint control.
    }
    \label{fig:tiago_single_joint_error}
\end{figure}
Figure~\ref{fig:tiago_single_joint_error} summarises the final absolute error 
across all seven Tiago arm joints. The controller successfully drives directed 
motion across all joints, confirming that the gain-modulated ring signal 
translates to physical displacement. Performance is joint-dependent: some joints 
show consistently small errors across batches, while others exhibit larger spread 
and occasional outliers. This variability reflects differences in joint range, 
mechanical dynamics, and the calibration of the neural drive to physical 
velocity, rather than failures of the control mechanism itself.

\subsection{NEST implementation of the forward-kinematic ring decoder}

We finally evaluate whether the product-feature readout derived in Section~2.5 
can be implemented in spiking neurons and decoded into orientation-coordinate 
output rings. We instantiate a two-ring decoder for a known two-revolute-joint 
chain,
\begin{equation}
    R(q_1, q_2) = R_x(q_1)\, R_y(q_2),
    \label{eq:two_ring_chain}
\end{equation}
where the first ring represents rotation about the \(x\)-axis and the second 
about the \(y\)-axis. Although only two input rings are used, the experiment 
tests the essential compositional mechanism: local \(SO(2)\) Fourier features 
are combined into signed product-feature populations via push-pull pairs, and 
those product features are projected to output rings encoding lift, pitch, and 
yaw.

\begin{figure}[t]
    \centering
    \includegraphics[width=\linewidth]{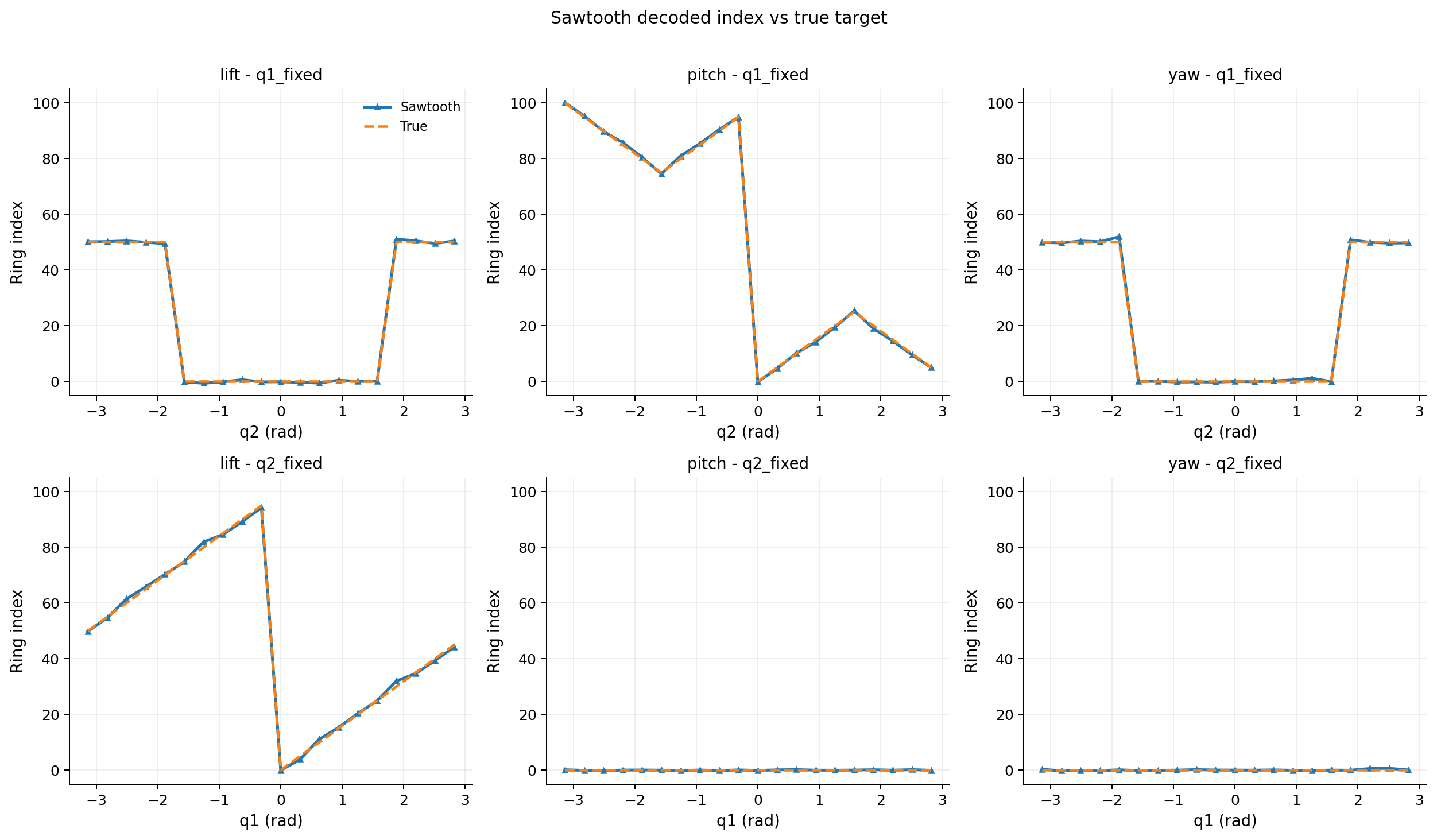}
    \caption{
    Sawtooth-decoded output-ring index against the true target index for two 
    slices of the joint configuration space (\(q_1\) fixed with \(q_2\) varied; 
    \(q_2\) fixed with \(q_1\) varied). Decoded and true traces closely overlap 
    for lift, pitch, and yaw. Discontinuities are expected sawtooth wrap-around 
    effects from the circular representation, quantified in 
    Figure~\ref{fig:multi_ring_circular_error}. See Appendix Figure~\ref{fig:multi_ring_decoder_profiles} for decoder activity.
    }
    \label{fig:multi_ring_sawtooth_validation}
\end{figure}

Figure~\ref{fig:multi_ring_decoder_profiles} shows the internal decoder 
activity. During a sweep of \(q_2\) with \(q_1\) fixed, the first-order 
features track the sweep directly while the product features reflect the 
interaction between the fixed \(q_1\) state and the varying \(q_2\) state. The 
output ring profiles confirm that these product features are sufficient to 
produce structured orientation-coordinate population responses.

Figure~\ref{fig:multi_ring_sawtooth_validation} quantifies the decoder output 
against analytically computed targets across two configuration-space slices. The 
decoded traces closely track the true targets throughout, and 
Figure~\ref{fig:multi_ring_circular_error} shows that the circular index error 
remains small across both sweep conditions and all three orientation variables. 
The discontinuities visible in Figure~\ref{fig:multi_ring_sawtooth_validation} 
are expected sawtooth wrap-around effects, not decoder failures.

These results validate the orientation readout in spiking neurons. The decoder 
does not control orientation --- it reads out the orientation implied by the 
ring states. The single-ring homeostatic gain-modulation loop generates 
\(SO(2)\) trajectories; the multi-ring product-feature decoder composes those 
trajectories into lift, pitch, and yaw output rings.

\section{Conclusion}

We introduced a spiking ring-attractor architecture for representing and
controlling revolute robot joints. Rather than treating joint angle as an
external scalar variable passed to a conventional controller, the proposed system
represents joint state as a persistent activity bump on a circular neural
manifold and regulates that state through homeostatic comparison and opponent
gain modulation. This provides a different paradigm for spiking robot control:
representation and action are not separated, but are coupled through transport of
a population code whose geometry matches the topology of the physical joint.

The results show that gain imbalance can induce directed bump motion on an
\(SO(2)\)-structured ring, that distributed ring activity can be decoded through
analytic Fourier features, and that local ring features can be composed into
forward-kinematic orientation readouts for known revolute chains. Robot
experiments further demonstrate that the gain-modulated ring dynamics can be
coupled to physical joint motion, producing directed actuation with
joint-dependent accuracy. Although the present implementation is not intended to
replace optimized model-based or PID controllers, it establishes a neuromorphic
control primitive in which circular state representation, regulation, and motor
drive are implemented within a common spiking substrate.

Because the controlled variable is represented by a population activity pattern
rather than a single scalar state, this framework may offer robustness advantages
under noise, partial perturbation, or neuronal variability: errors in individual
units need not destroy the represented phase as long as the population bump
remains coherent. More broadly, precise control of such attractor bumps could
enable applications beyond low-level joint actuation, including task-space
control, internal simulation, and neural forms of motor imagery. In biological
systems, structured activity patterns can be maintained, shifted, and replayed
offline, as in hippocampal experience replay; analogous mechanisms in robotic
ring-attractor substrates could support imagined trajectories, predictive
rollouts, or planning over embodied state spaces. We therefore view controlled
spiking attractor dynamics as a promising route toward natural neural controllers
for revolute systems, where the geometry of the controller is aligned with the
geometry of the body it controls.

\FloatBarrier

\bibliographystyle{plainnat}
\bibliography{references}

\clearpage
\appendix

\section{Ring-attractor mode analysis and SO(2) reduction}

This appendix provides the derivation underlying Section 2.2. We start from the continuum ring neural field
\[
    \tau_r \partial_t u_r(\theta,t)
    =
    -u_r(\theta,t)
    +
    \int_{-\pi}^{\pi}
    W(\theta-\theta')g(u_r(\theta',t))\,d\theta'
    +
    I(\theta,t),
\]
where \(u_r(\theta,t)\) is the ring activity, \(W(\theta-\theta')\) is a translation-invariant recurrent kernel, \(g\) is a static nonlinearity, and \(I\) is external input. For the mode analysis, we first consider a constant input \(I(\theta,t)=I_0\) and a spatially homogeneous equilibrium
\[
    u_r(\theta,t)=\bar{u}.
\]
Substituting this into the neural field and setting \(\partial_t u_r=0\) gives
\[
    \bar{u}
    =
    g(\bar{u})
    \int_{-\pi}^{\pi}
    W(\theta-\theta')\,d\theta'
    +
    I_0.
\]
Since the kernel depends only on the phase difference, the integral is independent of \(\theta\). Let
\[
    \widehat{W}_0
    =
    \int_{-\pi}^{\pi}
    W(\psi)\,d\psi .
\]
Then the homogeneous fixed point satisfies
\[
    \bar{u}
    =
    g(\bar{u})\widehat{W}_0 + I_0.
\]

To study the stability of this homogeneous state, write
\[
    u_r(\theta,t)=\bar{u}+v(\theta,t),
\]
where \(v\) is a small perturbation. Expanding the nonlinearity to first order,
\[
    g(\bar{u}+v)
    \approx
    g(\bar{u})+g'(\bar{u})v.
\]
Substituting into the neural field and cancelling the fixed-point terms gives the linearized dynamics
\[
    \tau_r \partial_t v(\theta,t)
    =
    -v(\theta,t)
    +
    g'(\bar{u})
    \int_{-\pi}^{\pi}
    W(\theta-\theta')v(\theta',t)\,d\theta'.
\]

Because the recurrent operator is convolutional on the ring, Fourier modes diagonalize the linearized dynamics. Let
\[
    v(\theta,t)
    =
    \sum_{k=-\infty}^{\infty}
    a_k(t)e^{ik\theta}.
\]
The Fourier coefficient of the kernel is
\[
    \widehat{W}_k
    =
    \int_{-\pi}^{\pi}
    W(\psi)e^{-ik\psi}\,d\psi.
\]
Using the convolution identity on the ring, each mode evolves independently:
\[
    \tau_r \dot{a}_k(t)
    =
    \left[-1+g'(\bar{u})\widehat{W}_k\right]a_k(t).
\]
Thus the growth rate of mode \(k\) is
\[
    \lambda_k
    =
    {1\over \tau_r}
    \left[-1+g'(\bar{u})\widehat{W}_k\right].
\]
Modes with \(\lambda_k<0\) decay, while modes with \(\lambda_k>0\) destabilize the homogeneous state. In the parameter regime considered in the main text, the recurrent kernel and nonlinearity support a stable single-bump profile rather than a homogeneous firing state. The mode analysis therefore motivates the reduction from a full activity field to a low-dimensional bump coordinate.

We assume the ring is operating in this activated single-bump regime. The activity can then be approximated by a fixed bump profile translated by a phase coordinate,
\[
    u_r(\theta,t)
    \approx
    f(\theta-\phi(t)),
\]
where \(f\) is the bump profile and \(\phi(t)\in S^1\) is the bump centroid. This is a collective-coordinate approximation: the controlled state is not the activity of a single neuron, but the position of the distributed bump.

The rotation action on ring activity is
\[
    (\rho_\alpha u_r)(\theta)=u_r(\theta-\alpha).
\]
For the bump state
\[
    u_\phi(\theta)=f(\theta-\phi),
\]
we have
\[
    (\rho_\alpha u_\phi)(\theta)
    =
    f(\theta-\alpha-\phi)
    =
    u_{\phi+\alpha}(\theta).
\]
Thus the family of translated bump states is an orbit of the continuous rotation group. The bump centroid may equivalently be represented by the rotation matrix
\[
    R(\phi)
    =
    \begin{pmatrix}
    \cos\phi & -\sin\phi \\
    \sin\phi & \cos\phi
    \end{pmatrix}
    \in SO(2).
\]
The full neural population state lives in a high-dimensional activity space, but the controlled collective coordinate \(\phi(t)\) lies on \(S^1\) and carries the action of \(SO(2)\).

The infinitesimal generator of rotations on the ring is
\[
    J=-\partial_\theta.
\]
For a small displacement \(\delta\phi\),
\[
    u_{\phi+\delta\phi}(\theta)
    =
    f(\theta-\phi-\delta\phi)
    \approx
    f(\theta-\phi)
    -
    \delta\phi \partial_\theta f(\theta-\phi).
\]
Using \(J=-\partial_\theta\), this becomes
\[
    u_{\phi+\delta\phi}(\theta)
    \approx
    u_\phi(\theta)+\delta\phi Ju_\phi(\theta).
\]
Thus the tangent direction of the single-bump manifold is \(Ju_\phi\). A control input that moves the bump should therefore project onto this direction.

\section{Derivation of gain modulation as a transport equation}

This appendix provides the derivation underlying Section 2.3. The goal is to show how opponent gain populations produce a first-order transport term on the ring. Let \(u_g^+(\theta,t)\) and \(u_g^-(\theta,t)\) denote the two gain populations. We model their inputs as shifted copies of the ring activity plus opponent homeostatic drive:
\[
    u_g^+(\theta,t)
    =
    W_{rg}g(u_r(\theta+\delta\theta,t))+B_+(t),
\]
\[
    u_g^-(\theta,t)
    =
    W_{rg}g(u_r(\theta-\delta\theta,t))+B_-(t),
\]
where \(W_{rg}\) is the ring-to-gain connection strength, \(\delta\theta\) is a small connection phase shift, and \(B_+(t)\), \(B_-(t)\) are homeostatic drive terms.

For small \(\delta\theta\), expand the shifted ring activity:
\[
    u_r(\theta+\delta\theta,t)
    \approx
    u_r(\theta,t)+\delta\theta\partial_\theta u_r(\theta,t),
\]
\[
    u_r(\theta-\delta\theta,t)
    \approx
    u_r(\theta,t)-\delta\theta\partial_\theta u_r(\theta,t).
\]
Applying the nonlinearity gives
\[
    g(u_r(\theta+\delta\theta,t))
    \approx
    g(u_r)
    +
    \delta\theta g'(u_r)\partial_\theta u_r,
\]
\[
    g(u_r(\theta-\delta\theta,t))
    \approx
    g(u_r)
    -
    \delta\theta g'(u_r)\partial_\theta u_r,
\]
where \(u_r\) abbreviates \(u_r(\theta,t)\).

Define the baseline gain inputs
\[
    x_+
    =
    W_{rg}g(u_r)+B_+(t),
    \qquad
    x_-
    =
    W_{rg}g(u_r)+B_-(t).
\]
Then the gain-population activities can be written to first order as
\[
    u_g^+
    \approx
    x_+
    +
    W_{rg}\delta\theta g'(u_r)\partial_\theta u_r,
\]
\[
    u_g^-
    \approx
    x_-
    -
    W_{rg}\delta\theta g'(u_r)\partial_\theta u_r.
\]
Expanding the gain output nonlinearity around \(x_+\) and \(x_-\) gives
\[
    g(u_g^+)
    \approx
    g(x_+)
    +
    W_{rg}\delta\theta g'(x_+)g'(u_r)\partial_\theta u_r,
\]
\[
    g(u_g^-)
    \approx
    g(x_-)
    -
    W_{rg}\delta\theta g'(x_-)g'(u_r)\partial_\theta u_r.
\]
Summing the two opponent outputs gives
\[
    g(u_g^+)+g(u_g^-)
    \approx
    g(x_+)+g(x_-)
    +
    W_{rg}\delta\theta g'(u_r)
    \left[g'(x_+)-g'(x_-)\right]
    \partial_\theta u_r.
\]
The first two terms are nondirectional. They modify the local drive to the ring but do not select a direction of motion. The final term is directional because it is proportional to \(\partial_\theta u_r\).

This expression also shows why balanced opponent drive does not move the bump. If \(B_+(t)=B_-(t)\), then \(x_+=x_-\), and therefore
\[
    g'(x_+)-g'(x_-)=0.
\]
The directional term vanishes. Thus equal opponent magnitudes preserve the rotational symmetry of the ring and do not generate bump transport. Directed motion requires \(B_+(t)\neq B_-(t)\), which breaks the symmetry and leaves a nonzero first-order transport term.

The state ring receives feedback from the gain populations with strength \(W_{gr}\). Substituting the directional term into the ring equation and collecting all nondirectional terms into \(F_r(u_r)\) gives
\[
    \partial_t u_r(\theta,t)
    +
    c(u_r)\partial_\theta u_r(\theta,t)
    =
    F_r(u_r).
\]
The symmetric part is
\[
    F_r(u_r)
    =
    {1\over \tau_r}
    \left[
    -u_r(\theta,t)
    +
    \int_{-\pi}^{\pi}
    W(\theta-\theta')g(u_r(\theta',t))\,d\theta'
    +
    I(\theta,t)
    \right].
\]
The transport coefficient is
\[
    c(u_r)
    =
    {W_{gr}W_{rg}\delta\theta\over \tau_r}
    g'(u_r)
    \left[g'(x_+)-g'(x_-)\right].
\]
Depending on the sign convention used for the feedback projection, the sign of \(c(u_r)\) may be absorbed into \(W_{gr}\) or into the definitions of the positive and negative gain channels. The important point is that the transport coefficient is proportional to the phase shift \(\delta\theta\), the bidirectional coupling \(W_{gr}W_{rg}\), and the local gain difference between the two opponent channels.

Using \(J=-\partial_\theta\), the transport equation becomes
\[
    \partial_t u_r(\theta,t)
    =
    F_r(u_r)
    +
    c(u_r)Ju_r(\theta,t).
\]
Thus gain modulation supplies the scalar coefficient multiplying the Lie algebra generator of the ring.

Finally, consider the single-bump approximation
\[
    u_r(\theta,t)=f(\theta-\phi(t)).
\]
Then
\[
    \partial_t u_r(\theta,t)
    =
    -\dot{\phi}(t)f'(\theta-\phi(t))
    =
    \dot{\phi}(t)Ju_r(\theta,t).
\]
If \(c(u_r)\) is approximately constant over the bump support, or if it is projected onto the bump tangent direction, the dynamics reduce to an effective phase equation
\[
    \dot{\phi}(t)=c_{\mathrm{eff}}(t),
\]
where \(c_{\mathrm{eff}}(t)\) is the transport coefficient averaged or projected over the bump. Therefore opponent gain modulation encodes velocity along the \(SO(2)\)-structured bump manifold, rather than encoding the joint angle directly.

\section{Fourier decoder derivation}

This appendix provides the derivation underlying the positional Fourier decoder in Section 2.4. The problem is to convert a distributed bump state into a scalar phase estimate that can be used by the homeostatic comparator.

Let the ring activity be a translated bump profile
\[
    g(x)=r(x-\theta),
\]
where \(x=0,\ldots,N\) indexes discrete ring positions and \(\theta\) is the bump position. The centroid estimator is
\[
    \hat{\theta}_{\mathrm{cent}}
    =
    {\sum_{x=0}^{N}xg(x)\over \sum_{x=0}^{N}g(x)}.
\]
Assume that the bump shape and amplitude are approximately constant under translation. Then the total mass
\[
    M_0
    =
    \sum_{x=0}^{N}g(x)
\]
is approximately independent of \(\theta\). The centroid can then be written as the linear decoder
\[
    \hat{\theta}
    \approx
    {1\over M_0}
    \sum_{x=0}^{N}xg(x).
\]

To center the decoded coordinate, define
\[
    L=N+1,
    \qquad
    \xi(x)=x-{L\over 2}.
\]
Then
\[
    \hat{\theta}-{L\over 2}
    \approx
    {1\over M_0}
    \sum_{x=0}^{N}\xi(x)g(x).
\]
Thus position decoding is equivalent to projecting the activity profile onto a centered ramp function.

On a periodic interval, the centered ramp is represented by a sine series. For \(x\in[0,L]\),
\[
    x-{L\over 2}
    =
    -{L\over \pi}
    \sum_{m=1}^{\infty}
    {1\over m}
    \sin\left({2\pi mx\over L}\right).
\]
Substituting this into the centered centroid estimator gives
\[
    \hat{\theta}-{L\over 2}
    \approx
    {1\over M_0}
    \sum_{x=0}^{N}
    g(x)
    \left[
    -{L\over \pi}
    \sum_{m=1}^{\infty}
    {1\over m}
    \sin\left({2\pi mx\over L}\right)
    \right].
\]
Interchanging the order of summation yields
\[
    \hat{\theta}-{L\over 2}
    \approx
    -{L\over \pi M_0}
    \sum_{m=1}^{\infty}
    {1\over m}
    \sum_{x=0}^{N}
    g(x)
    \sin\left({2\pi mx\over L}\right).
\]
Define the \(m\)-th sine feature
\[
    S_m(g)
    =
    \sum_{x=0}^{N}
    g(x)
    \sin\left({2\pi mx\over L}\right).
\]
Then
\[
    \hat{\theta}
    \approx
    {L\over 2}
    -
    {L\over \pi M_0}
    \sum_{m=1}^{\infty}
    {1\over m}S_m(g).
\]
Truncating the series to \(K\) modes gives the practical decoder
\[
    \hat{\theta}
    =
    b+\sum_{m=1}^{K}w_mS_m(g),
\]
with
\[
    b={L\over 2},
    \qquad
    w_m=-{L\over \pi M_0m}.
\]

If the bump profile has finite width, then each sine feature is attenuated by the Fourier response of the bump shape. In this case,
\[
    S_m(g)
    \approx
    A_m\sin\left({2\pi m\theta\over L}\right),
\]
where \(A_m\) depends on the shape and width of the bump. A shape-corrected decoder can therefore absorb this attenuation into the weights:
\[
    w_m=-{L\over \pi m A_m}.
\]
This shows that the Fourier decoder is a periodic reformulation of the centroid estimator. The centroid projects activity onto a ramp; the Fourier decoder expresses that ramp in a sine basis adapted to the ring geometry.

\section{Signed implementation of Fourier features}

The sine features used by the decoder can take positive or negative values, whereas firing rates and spike counts are nonnegative. To implement signed Fourier features with nonnegative neural activity, each sine feature is represented by two subpopulations:
\[
    S_m^+(x)=\max\left(\sin\left({2\pi mx\over L}\right),0\right),
\]
\[
    S_m^-(x)=\max\left(-\sin\left({2\pi mx\over L}\right),0\right).
\]
The signed feature is recovered by subtraction:
\[
    S_m(g)=S_m^+(g)-S_m^-(g).
\]
In practice, the two subpopulations accumulate spike responses to the positive and negative parts of the sine feature. The downstream linear decoder then combines the recovered signed features to estimate the bump phase. This implements the Fourier phase readout using nonnegative neural responses while preserving the linear decoding structure used in the main text.

\section{Compositional \(SO(2)\) ring decoder for \(SO(3)\) orientation}

This appendix derives the orientation readout used in the robot experiments. 
The control mechanism --- homeostatic comparison driving opponent gain modulation 
on individual rings --- is described in the main text. Here we show how the 
decoded phases of several such rings compose into a robot-level \(SO(3)\) 
orientation estimate through forward kinematics. This is a readout, not an 
additional control law.

\subsection{From ring activity to local trigonometric features}

The Fourier readout derived in Appendix~C extracts a scalar phase estimate from 
each ring. For the orientation readout we retain the trigonometric form rather 
than collapsing to a scalar angle, because forward kinematics is naturally 
expressed in terms of sines and cosines. For ring \(i\) with activity profile 
\(g_i(x)\) over positions \(x = 0,\ldots,L-1\), the first harmonic Fourier 
features are
\begin{equation}
    C_i = \sum_{x=0}^{L-1} g_i(x)\cos\!\left(\frac{2\pi x}{L}\right),
    \qquad
    S_i = \sum_{x=0}^{L-1} g_i(x)\sin\!\left(\frac{2\pi x}{L}\right).
    \label{eq:fourier_features}
\end{equation}
When bump shape and amplitude are approximately preserved under translation --- 
the same condition that justifies the centroid approximation in the main text --- 
normalising by the bump shape gives reliable estimates
\begin{equation}
    \bar{C}_i \approx \cos q_i,
    \qquad
    \bar{S}_i \approx \sin q_i,
    \label{eq:normalised_features}
\end{equation}
where \(q_i\) is the phase represented by the bump centroid. These are collected 
into a local feature vector
\begin{equation}
    \varphi_i
    =
    \begin{bmatrix} 1 \\ \bar{C}_i \\ \bar{S}_i \end{bmatrix}
    \approx
    \begin{bmatrix} 1 \\ \cos q_i \\ \sin q_i \end{bmatrix},
    \label{eq:local_feature_vec}
\end{equation}
where the constant term accounts for the fixed entries present in all 
coordinate-axis rotation matrices.

\subsection{A single revolute rotation is linear in ring features}

A rotation about a coordinate axis \(a_i \in \{x, y, z\}\) takes the form
{\setlength{\arraycolsep}{2.5pt}
\begin{equation}
    R_x(q_i)
    =
    \begin{bmatrix}
        1 & 0 & 0 \\
        0 & \cos q_i & -\sin q_i \\
        0 & \sin q_i & \cos q_i
    \end{bmatrix},
    \;
    R_y(q_i)
    =
    \begin{bmatrix}
        \cos q_i & 0 & \sin q_i \\
        0 & 1 & 0 \\
        -\sin q_i & 0 & \cos q_i
    \end{bmatrix},
    \;
    R_z(q_i)
    =
    \begin{bmatrix}
        \cos q_i & -\sin q_i & 0 \\
        \sin q_i & \cos q_i & 0 \\
        0 & 0 & 1
    \end{bmatrix}.
    \label{eq:coord_rotations}
\end{equation}}
Each matrix is linear in \(1, \cos q_i, \sin q_i\), so for any coordinate axis 
the local rotation can be written as
\begin{equation}
    R_{a_i}(q_i)
    =
    B^{(0)}_{a_i}
    +
    \cos q_i\, B^{(c)}_{a_i}
    +
    \sin q_i\, B^{(s)}_{a_i},
    \label{eq:linear_local_rot}
\end{equation}
where the coefficient matrices \(B^{(0)}, B^{(c)}, B^{(s)}\) are read off 
directly from equation~\eqref{eq:coord_rotations} and are fixed once the joint 
axis is known. A single ring's Fourier features therefore suffice to linearly 
represent one coordinate-axis revolute rotation.

\subsection{Composing rings produces product Fourier features}

For a serial chain of \(n\) revolute joints, the end-effector orientation is the 
ordered product
\begin{equation}
    R_{\mathrm{ee}}(q_1,\ldots,q_n)
    =
    R_{a_1}(q_1)\, R_{a_2}(q_2) \cdots R_{a_n}(q_n).
    \label{eq:fk_product}
\end{equation}
The order here is not arbitrary --- serial rotations do not commute, so the 
multiplication order is part of the kinematic convention and must match the robot 
model. Substituting equation~\eqref{eq:linear_local_rot} for each factor and 
expanding yields matrix entries that are sums of products of local sine and 
cosine terms across joints, such as
\begin{equation}
    \cos q_1 \cos q_2, \qquad
    \cos q_1 \sin q_2, \qquad
    \sin q_1 \sin q_2, \qquad
    \cos q_1 \cos q_2 \cos q_3.
    \label{eq:cross_terms}
\end{equation}
All such terms appear in the tensor product of the individual feature vectors. 
The chain feature vector
\begin{equation}
    \Phi^{(n)}
    =
    \varphi_1 \otimes \varphi_2 \otimes \cdots \otimes \varphi_n
    \label{eq:chain_features_app}
\end{equation}
contains every cross-term generated by the kinematic chain and has \(3^n\) 
entries for \(n\) first-harmonic rings. Since every entry of 
\(R_{\mathrm{ee}}\) is a sum of these product features, there exists a fixed 
matrix \(T\) such that
\begin{equation}
    \mathrm{vec}(R_{\mathrm{ee}}) = T\Phi^{(n)},
    \label{eq:exact_readout}
\end{equation}
which is exact for coordinate-axis revolute chains. For the three-ring case 
considered here \(\Phi^{(n)}\) has \(3^3 = 27\) entries, which is entirely 
manageable; longer chains would require feature pruning or low-rank 
approximation.

\subsection{Neural implementation with estimated features}

In the spiking model the exact features are replaced by ring-activity estimates, 
giving
\begin{equation}
    \hat{\Phi}^{(n)}
    =
    \hat{\varphi}_1 \otimes \hat{\varphi}_2 \otimes \cdots \otimes \hat{\varphi}_n,
    \qquad
    \mathrm{vec}(\hat{R}_{\mathrm{ee}}) = T\hat{\Phi}^{(n)}.
    \label{eq:estimated_readout}
\end{equation}
Because product Fourier features can take positive or negative values while 
neural firing rates are nonnegative, each feature is implemented via a push-pull 
pair of subpopulations,
\begin{equation}
    \Phi^+_j = \max(\Phi_j, 0),
    \qquad
    \Phi^-_j = \max(-\Phi_j, 0),
    \qquad
    \Phi_j = \Phi^+_j - \Phi^-_j,
    \label{eq:signed_impl}
\end{equation}
preserving the linear readout structure while remaining compatible with 
nonnegative spike counts. The signed product populations project directly to 
output rings for lift, pitch, and yaw using analytic or learned weights.

\subsection{Three-ring orientation readout for lift, pitch, and yaw}

For the robot experiments three ring coordinates represent lift, pitch, and yaw. 
Under the ordered convention
\begin{equation}
    R = R_{\mathrm{lift}}(\hat{q}_{\mathrm{lift}})\,
        R_{\mathrm{pitch}}(\hat{q}_{\mathrm{pitch}})\,
        R_{\mathrm{yaw}}(\hat{q}_{\mathrm{yaw}}),
    \label{eq:euler_convention}
\end{equation}
the composed feature vector \(\Phi = \varphi_{\mathrm{lift}} \otimes 
\varphi_{\mathrm{pitch}} \otimes \varphi_{\mathrm{yaw}}\) supports the exact 
linear readout \(\mathrm{vec}(R) = T\Phi\). When an explicit rotation matrix is 
decoded, lift, pitch, and yaw are recovered as
\begin{equation}
    \hat{\theta} = -\arcsin(r_{31}),
    \qquad
    \hat{\lambda} = \arctan2(r_{32}, r_{33}),
    \qquad
    \hat{\psi} = \arctan2(r_{21}, r_{11}),
    \label{eq:euler_extraction}
\end{equation}
valid away from the gimbal singularity at \(\cos\theta = 0\), which does not 
arise in the workspace used for the Tiago experiments. Alternatively, a direct 
linear decoder from \(\hat{\Phi}\) to each output ring coordinate can be fitted, 
bypassing explicit matrix extraction and allowing the composition to be carried 
out entirely within the neural substrate.

The \(SO(3)\) orientation produced by this readout is not separately controlled. 
It is the forward-kinematic image of three independently controlled \(SO(2)\) 
ring trajectories, each driven by its own homeostatic gain-modulation loop.

\newpage
\section{Extra Results}

\subsection{Tiago single-joint control from gain-modulated ring velocity}
\begin{figure}[!htbp]
    \centering
    \includegraphics[width=\linewidth]{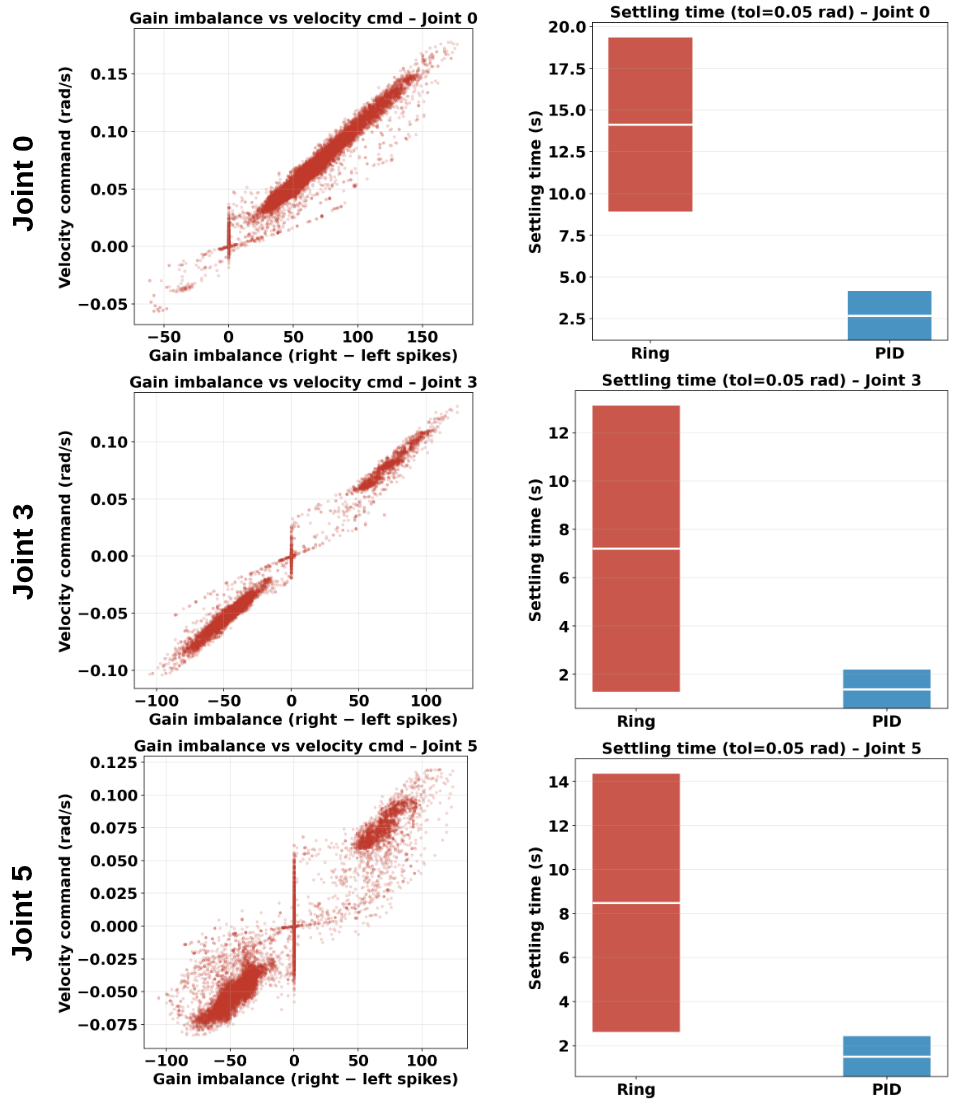}
    \caption{
    Left: Gain-population imbalance versus decoded velocity command across joints 0,3 and 5. Each point represents one recorded time step from the ring controller, showing an approximately linear increase in decoded velocity with the difference between right and left gain-population spike counts.
    Right: Settling time for joints 0, 3, and 5 under the ring and PID controllers. Settling time is defined as the first time the absolute tracking error remains below 0.05 rad for 5 consecutive samples; bars show mean ± 1 SD across trials.
    }
    \label{fig:single_joint_extra_info}
\end{figure}

\begin{figure}[!htbp]
    \centering
    \includegraphics[width=\linewidth]{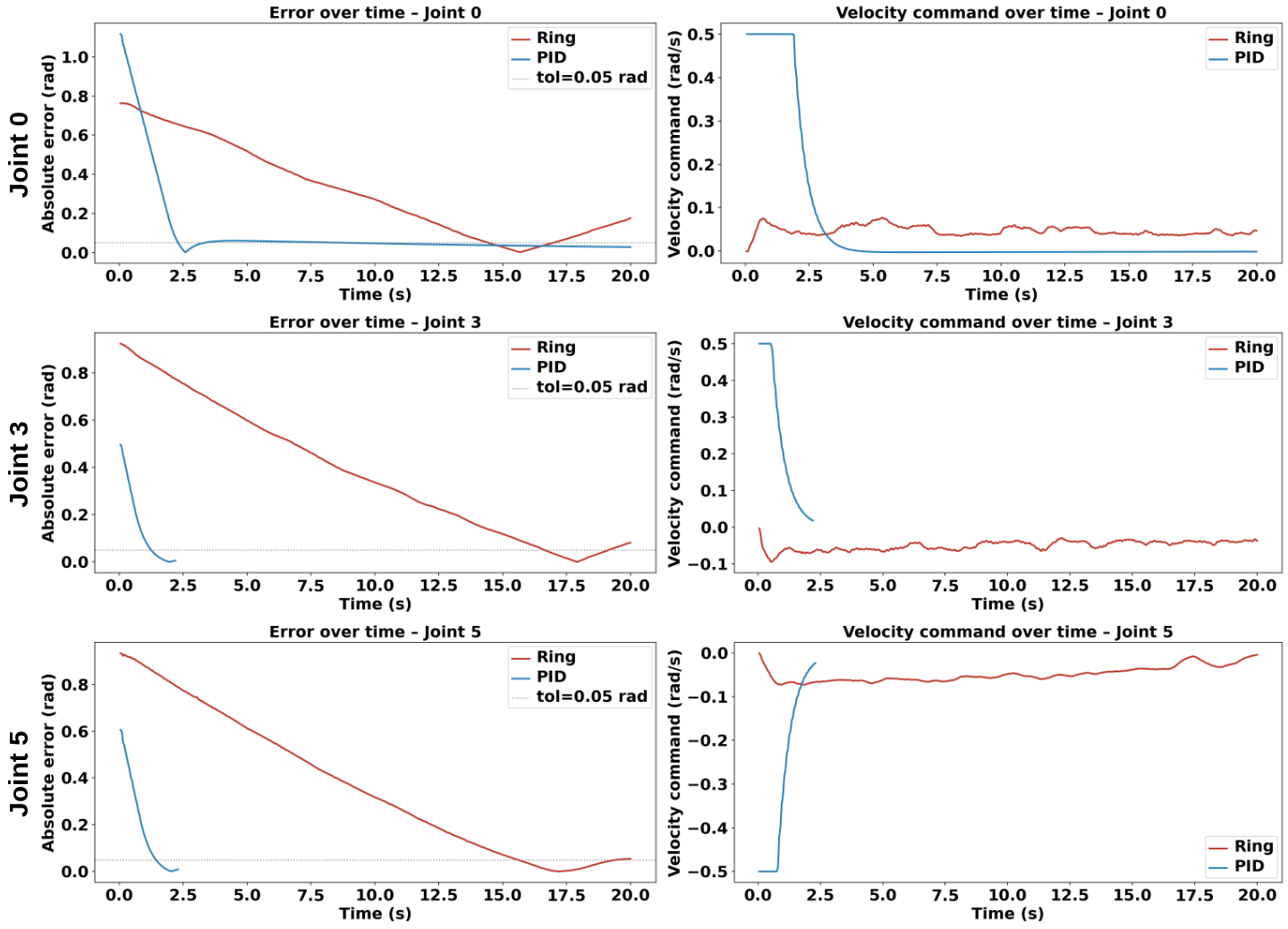}
    \caption{
    Left: Absolute tracking error over time for joints 0, 3, and 5, shown for representative Ring and PID trials. The dashed line indicates the 0.05 rad tolerance. PID drops into this band quickly, while the Ring controller reduces error more slowly and reaches the same threshold later.
    Right: Velocity command over time for joints 0, 3, and 5, shown for representative Ring and PID trials. PID starts with a strong corrective command and rapidly decays toward zero, while the Ring controller applies smaller, smoother commands that vary gradually over a longer duration.
    }
    \label{fig:single_joint_extra_info_2}
\end{figure}

\clearpage
\subsection{NEST implementation of the forward-kinematic ring decoder}
\begin{figure}[!htbp]
    \centering
    \includegraphics[width=\linewidth]{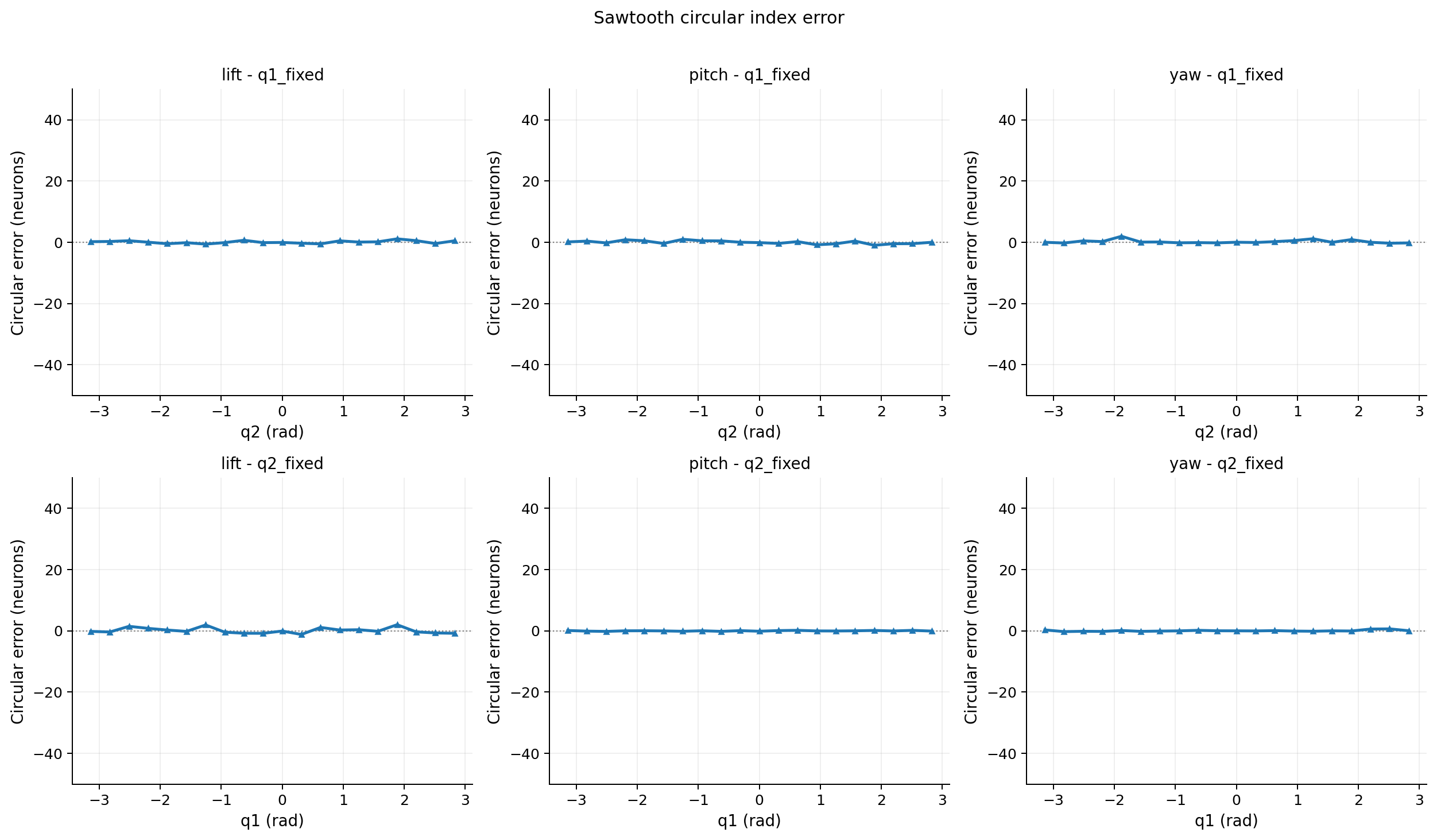}
    \caption{
    Circular index error for the two-ring forward-kinematic decoder under the 
    same sweep conditions as Figure~\ref{fig:multi_ring_sawtooth_validation}. 
    Across both sweeps and all three decoded orientation variables, the circular 
    error remains small, confirming that the discontinuities in 
    Figure~\ref{fig:multi_ring_sawtooth_validation} arise from the circular 
    representation rather than decoder failure.
    }
    \label{fig:multi_ring_circular_error}
\end{figure}

\begin{figure}[!htbp]
    \centering
    \includegraphics[width=\linewidth]{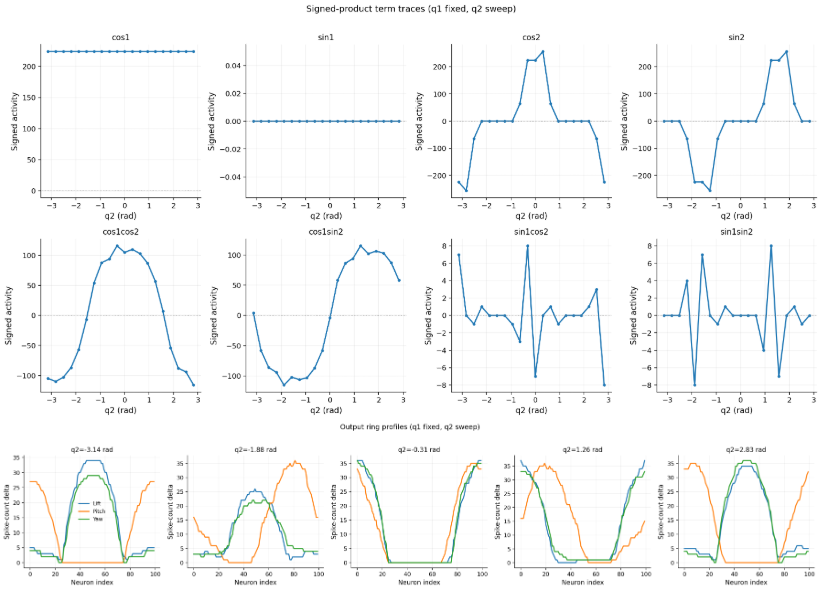}
    \caption{
    Spiking implementation of the two-ring forward-kinematic decoder in NEST. 
    Top: signed product-feature population activity during a sweep of \(q_2\) 
    with \(q_1\) fixed, showing systematic variation of first-order features 
    (\(\cos q_1, \sin q_1, \cos q_2, \sin q_2\)) and second-order products 
    (\(\cos q_1 \cos q_2\), etc.) with joint angle. Bottom: output ring profiles 
    for lift, pitch, and yaw at representative \(q_2\) values, showing structured 
    population responses driven by the product-feature layer.
    }
    \label{fig:multi_ring_decoder_profiles}
\end{figure}

\FloatBarrier
\newpage
\clearpage
\section{Model Parameters}

\begin{table}[!htbp]
\centering
\caption{Spiking model parameters used for the ring, homeostatic comparator, and gain-modulation populations.}
\label{tab:model_params}
\small
\setlength{\tabcolsep}{5pt}
\renewcommand{\arraystretch}{1.12}
\begin{tabular}{@{}llccc@{}}
\toprule
\textbf{Parameter} & \textbf{Symbol} & \textbf{Ring} & \textbf{Homeostatic} & \textbf{Gain} \\
\midrule
\multicolumn{5}{@{}l}{\textit{Neuron parameters for \texttt{iaf\_psc\_alpha}}} \\
Firing threshold (mV) 
    & $V_{\mathrm{th}}$ 
    & 0.8 & 0.8 & 1.35 \\
Membrane time constant (ms) 
    & $\tau_m$ 
    & 7.04 & 7.04 & 7.04 \\
Membrane capacitance (pF) 
    & $C_m$ 
    & $3.96{\times}10^6$ & $3.96{\times}10^6$ & $3.96{\times}10^6$ \\
Excitatory synaptic time constant (ms) 
    & $\tau_{\mathrm{syn}}^{\mathrm{ex}}$ 
    & 27.0 & 75.0 & 27.0 \\
Inhibitory synaptic time constant (ms) 
    & $\tau_{\mathrm{syn}}^{\mathrm{in}}$ 
    & 50.0 & 50.0 & 50.0 \\
Background current (pA) 
    & $I_e$ 
    & $4.5{\times}10^5$ & $9.0{\times}10^5$ & 0 \\
Refractory period (ms) 
    & $t_{\mathrm{ref}}$ 
    & 0 & 0 & 0 \\
\midrule
\multicolumn{5}{@{}l}{\textit{Ring-attractor topology and readout}} \\
Population size 
    & $N$ 
    & \multicolumn{3}{c}{200} \\
Represented positions 
    & $N_{\mathrm{pos}}$ 
    & \multicolumn{3}{c}{30} \\
Readout weight scale 
    & $s_{\mathrm{readout}}$ 
    & \multicolumn{3}{c}{200} \\
Ridge regularization 
    & $\lambda$ 
    & \multicolumn{3}{c}{$10^{-6}$} \\
\midrule
\multicolumn{5}{@{}l}{\textit{Gain-modulation connectivity}} \\
Homeostatic comparator to gain 
    & $w_{h \to g}$ 
    & \multicolumn{3}{c}{$1.0{\times}10^4$} \\
Ring to gain 
    & $w_{r \to g}$ 
    & \multicolumn{3}{c}{$4.5{\times}10^5$} \\
Gain to ring 
    & $w_{g \to r}$ 
    & \multicolumn{3}{c}{$-0.6$} \\
Cross-inhibition 
    & $w_{\mathrm{cross}}$ 
    & \multicolumn{3}{c}{$-3.0{\times}10^4$} \\
\bottomrule
\end{tabular}
\end{table}

\end{document}